%% file: root.tex
\documentclass{ifacconf}

\usepackage{graphicx}      
\usepackage{natbib}        
\usepackage{amsmath}
\usepackage{amsfonts}
\usepackage{url}
\usepackage{booktabs}

\newcommand{\N}{\mathcal{N}}
\newcommand{\D}{\mathcal{D}}

\newcommand{\hp}{\phi}
\newcommand{\grs}{J}
\newcommand{\gr}{j}

\newcommand{\step}{i}
\newcommand{\steps}{T}

\newcommand{\R}{\mathbb{R}}

\newcommand{\seq}[1]{\mathbf{#1}}
\newcommand{\dt}{\frac{\mathrm{d}}{\mathrm{d}t}}

\newtheorem{remark}{Remark}

\begin{document}
\begin{frontmatter}

\title{Learning Dynamical Systems from Multiple Sparse Datasets: A Hierarchical Bayesian Modeling Approach}


\author[IDSIA]{Cristian Brugnara} 
\author[IDSIA]{Lea Multerer} 
\author[IDSIA]{Marco Forgione}
\author[IDSIA]{Laura Azzimonti}

\address[IDSIA]{SUPSI, Dalle Molle Institute for Artificial Intelligence (IDSIA USI-SUPSI), Lugano, Switzerland (e-mail: lea.multerer@idsia.ch)}

\begin{abstract}
\input{Content/00_Abstract}
\end{abstract}

\begin{keyword}
System identification,
Hierarchical Bayesian modeling,
Probabilistic meta-learning,
Sparse data,
Non-identifiability.
\end{keyword}

\end{frontmatter}

\input{Content/10_Introduction}
\input{Content/20_ProblemFormulation}
\input{Content/21_MetaLearning}
\input{Content/22_Implementation}
\input{Content/30_Experiment}
\input{Content/40_Conclusion}



\bibliography{Bibtex_Bayesian_Hierarchical_ODEs}

\end{document}

%% file: Content/00_Abstract.tex
Estimating parameters of dynamical systems from sparse, noisy, and irregularly sampled data is often severely ill-conditioned.
When multiple related datasets are available, they provide additional information if the shared structure and variability are properly modeled.
We propose a hierarchical Bayesian framework for probabilistic meta-learning in dynamical systems, modeling dataset-specific parameters as draws from a shared population distribution.
A numerical ODE solver is embedded within gradient-based MCMC to enable efficient posterior inference of the shared population and dataset-specific parameter distribution.
Experiments show improved predictive performance over unpooled methods, highlighting the potential for data-efficient system identification in settings with sparse data.

%% file: Content/10_Introduction.tex
\section{Introduction}

Gray-box models based on ordinary differential equations (ODEs) are a
standard description of time-evolving physical systems across domains
such as engineering, biology and epidemiology. In many applications, their
parameters must be estimated from small, noisy and irregularly sampled datasets.
When data informativity is extremely low, this inverse problem may be ill-conditioned, rendering the estimates unreliable or even undefined~\citep{gevers_identification_2009}. This is a manifestation of practical, as opposed to structural, non-identifiability~\citep{raue_structural_2009}.

In system identification, the classical remedy for low data informativity is regularization: prior information is injected to stabilize the estimate, either through an explicit penalty or through a tunable kernel whose hyperparameters are adjusted from the data by marginal-likelihood maximization~\citep{pillonetto_regularized_2022}. A complementary approach is Bayesian inference, where a probabilistic prior over parameters is updated by conditioning on the data to yield a full posterior over the unknowns.

A second lever appears when a \emph{collection} of experiments generated by
\emph{similar} dynamical systems is available; its usefulness is governed by how
much such systems vary. Assume all systems are identical and the collection reduces
to one long, likely informative experiment; assume them completely independent
and the shared structure is discarded, leaving each weak dataset to fend for
itself. Often, neither extreme fits: systems are related but
not identical, and the variability between them is itself unknown, and must be
inferred across the entire collection of datasets.

Estimating the across-dataset variability is precisely what mixed-effects, or
multilevel, modeling does, a construction with a long tradition in applied
statistics~\citep{pinheiro_mixed-effects_2000}. Its Bayesian instantiation, which we adopt
in this work, is known as Hierarchical Bayesian Modeling (HBM)~\citep{gelman_bayesian_2014}. In HBM, the dataset-specific parameters are treated
as draws from a shared distribution whose hyperparameters, including the spread,
are inferred jointly with the parameters. Sparse longitudinal data, as in
epidemiology~\citep{congdon_bayesian_2019, lawson_bayesian_2018}, and population
pharmacokinetics~\citep{wakefield_bayesian_1996} are among the most widespread
applications.

A limitation of current HBM implementations is that they require the
parameter-to-observation map to be linear or, at least, available in closed
form, which significantly restricts applicability. For dynamical
systems, in particular, this map runs through the ODE solution,
which generally admits no closed form in the nonlinear case, and can therefore only be evaluated approximately
through a numerical integration scheme.

As a result,
Bayesian inference for ODEs is largely limited to the
non-hierarchical single-dataset case~\citep{stuart_inverse_2010,
roda_bayesian_2020, green_bayesian_2015}.
Hierarchical multi-dataset extensions only exist for special cases where
analytical ODE solutions are known, reducing to nonlinear mixed-effects
regression~\citep{multerer_analysis_2021}, or where the ODE is replaced by
basis-function or collocation surrogates~\citep{loos_hierarchical_2018,
huang_bayesian_2020}. Approaches that both pool across related datasets and
support a numerical ODE solver 
within a Bayesian inference scheme 
remain rare.

A related line of work addresses low data informativity through 
meta-learning~\citep{forgione_system_2023, chakrabarty_meta_2025, 
lakshminarayanan_fine-tuning_2025}. By exploiting shared structure across a 
population of systems, meta-learning compensates for individually weak datasets, 
allowing the identification of even high-dimensional black-box models (e.g., neural networks). 
However, current approaches require a very large collection of related 
datasets, which in practice is feasible only when a digital twin of the process 
can synthesize them at scale. 

In this work, we instead target the regime of a 
modest number of genuinely measured datasets, each individually weak.
We formalize probabilistic meta-learning for gray-box ODE systems within a HBM framework, with
the ODE coefficients and noise parameters of each dataset forming the
dataset-specific level of the hierarchy. Approximate posterior inference is
carried out by Markov Chain Monte Carlo (MCMC) sampling.
Because the posterior dimension
grows with the number of datasets through the dataset-specific parameters,
gradient-free samplers become impractical; we therefore adopt the No-U-Turn Sampler (NUTS)~\citep{hoffman_no-u-turn_2014}, a gradient-based sampler that scales to
high-dimensional probabilistic problems. The enabling ingredient is a differentiable ODE
solver: embedding it in the computational graph allows gradients of the
likelihood to propagate through the numerical integration step, making
NUTS viable in the ODE setting. On a Lotka--Volterra benchmark, we show that systematic pooling across sparse,
noisy, and irregularly sampled trajectories markedly improves practical
identifiability and predictive accuracy over unpooled Bayesian and nonlinear
least-squares baselines.

The rest of this paper is organized as follows.
We introduce the problem formulation of meta-datasets with a shared structure in Section~\ref{sec:problem_formulation}.
In Section~\ref{sec:prob_meta_learning}, we develop the meta-learning approach within the HBM framework.
We detail the implementation in Section~\ref{sec:implementation} and present a numerical example in Section~\ref{sec:experiment}.
Finally, conclusions and directions for further work are discussed in Section~\ref{sec:conclusions}.

%% file: Content/20_ProblemFormulation.tex
\section{Problem Formulation}\label{sec:problem_formulation}

\subsection{System Description}
We consider dynamical systems described by a system of first-order ODEs:
\begin{subequations}
\label{eq:system_equation}
\begin{equation}
\label{eq:state_equation}
\dt x(t) = f\big(x(t), t;\; \theta_f\big), \qquad x(0) = x_0,
\end{equation}
where $x(t) \in \mathbb{R}^{n_x}$ denotes the state vector evolving over continuous time $t \in \mathbb{R}^+$, $f: \mathbb{R}^{n_x} \to \mathbb{R}^{n_x}$ defines the state dynamics, and $\theta_f \in \mathbb{R}^{n_{\theta_f}}$ are the uncertain parameters of $f$, which vary from system to system.

The systems are partially observed at a finite set of $\steps$ measurement instants irregularly spaced over time. Let $\seq{t} = \{t_1, t_2, \ldots, t_{\steps}\}$, $t_\step \in \mathbb{R}^+$ and $\seq{y} = \{y_1, y_2, \ldots, y_{\steps}\}$, $y_\step \in \R^{n_y}$ denote the set of sampling times and corresponding output measurements, respectively. The measurement $y_i\in \mathbb{R}^{n_y}$ at time step $t_i$ is: 
\begin{equation}
\label{eq:noise_on_state_equation}
    y(t_i) =  g\big(x(t_i);\;\theta_g\big) + \eta_i,
\end{equation}
\end{subequations}
where the function $g : \mathbb{R}^{n_x} \to \mathbb{R}^{n_y}$ has uncertain parameters $\theta_g \in \mathbb{R}^{n_{\theta_g}}$, also varying across system instances. The functional forms of $f$ and $g$ are assumed to be known.
Furthermore, $\eta_i\in \mathbb{R}^{n_y}$ is additive measurement noise generated by a probability density function $p_\eta(\cdot, \theta_\eta)$ that depends on further, potentially unknown, parameters $\theta_\eta \in \R^{n_{\theta_\eta}}$.
In the following, we denote the complete set of a system's uncertain parameters to be estimated
by $\theta = [\theta_f, \theta_g, \theta_\eta]^\top \in \R^{n_\theta}$.
\begin{remark}[Unknown Initial States]
In this work, we assume the initial state $x_0$ to be known. Handling an uncertain $x_0$ can be incorporated with only minor modifications.
\end{remark}
\begin{remark}[Non-Autonomous Systems]
We present results for \emph{autonomous} systems for notational compactness. The extension to the \emph{non-autonomous} case,
where the system depends on an exogenous input signal $u(t) \in \R^{n_u}$, 
is straightforward: $f(x, t;\;\theta_f)$ in \eqref{eq:state_equation} is simply replaced with $f\big(x, u(t), t;\; \theta_f\big)$.
\end{remark}

\subsection{Meta-Dataset and Shared Structure}
\label{sec:shared_structure}
A finite collection of $\grs$ datasets is available. All datasets are generated by systems of the defined ODE structure~\eqref{eq:system_equation}. However, each system is characterized by a different realization of the parameters $\theta$ and a potentially different set of measurement instants. In line with the meta-learning literature, this collection is referred to as the \emph{meta-dataset}:
\begin{subequations}
\begin{equation}
    \mathcal{D} = \{D^{1}, D^{2}, \ldots, D^{\grs}\},
\end{equation}
where the $\gr$-th dataset $D^\gr \in \D$ is characterized by $D^\gr = \{\mathbf{t}^\gr, \mathbf{y}^\gr\}$, with $\mathbf{t}^\gr = \{t_1^\gr, \dots, t_{T_\gr}^\gr\}$ and $\mathbf{y}^\gr = \{{y}_1^\gr, \dots, {y}_{T_\gr}^\gr\}$.

Some (or all) datasets $D^\gr \in \D$ are assumed to be individually \emph{information-poor}. This means that applying standard supervised learning techniques (e.g., maximum likelihood, maximum a posteriori, or full Bayesian inference under a weak prior) to a single dataset would yield unacceptably large uncertainty, or even an ill-posed estimation problem.

However, the meta-dataset exhibits a strong, yet a priori unknown \emph{similarity pattern}. Specifically, an unknown distribution $p^o(\theta)$ characterizes the generation of $\theta^j$ and, therefore, the common structure across datasets. We posit that, if $p^o(\theta)$ were known, the parameter posterior $p(\theta^{\gr} 
\mid D^{\gr};\; p^o)$ over each individual dataset, formally defined by:
\begin{equation}
\label{eq:exact_post_inference}
p(\theta^{\gr} 
\mid D^{\gr};\; p^o)
= \frac{p(D^{\gr} \mid \theta^{\gr})\, p^o(\theta^\gr)}{p(D^\gr)}
\end{equation}
\end{subequations}
would be sufficiently concentrated around the true $\theta^\gr$ to support reliable inference, for all $\gr = 1, 2, \dots, \grs$.

\subsection*{Example: LTI Meta-Dataset}
Consider a meta-dataset $\D$ generated by first-order linear-time-invariant (LTI) dynamical systems:
\begin{subequations}  
\label{eq:lti_structure}
\begin{equation}
    \dt x(t) = -\frac{1}{\tau}\big(x(t) - d \big),
\end{equation}
where $\tau$ is the time constant, $d$ is the steady-state value
and $x_0 = 0$ is the initial state, assumed known.
Note that in this simple example, the ODE solution $x(t) = d + (x_0 - d)\,e^{-t/\tau}$ is available in closed-form.
The measurement $y$ corresponds to the state variable $x$ corrupted by white Gaussian noise:
\begin{equation}
    y(t_\step) = x(t_\step) + \eta_\step,\qquad  \eta_\step \sim \N(0, \sigma_\eta^2),
\end{equation}
\end{subequations}
where the noise standard deviation ($\sigma_\eta = 0.05$) is known and fixed in all datasets.

Instead, the coefficients $\theta = [\tau \; d]^\top$ vary across datasets and are characterized by the (unknown) distribution $p^o(\theta)$:
\begin{equation}    
\label{eq:lti_generative}
\tau \sim U(0.9, 1.1), \qquad d \sim U(0.5, 2.0).
\end{equation}

\begin{figure}[htbp]
\centering
\includegraphics[width=\columnwidth]{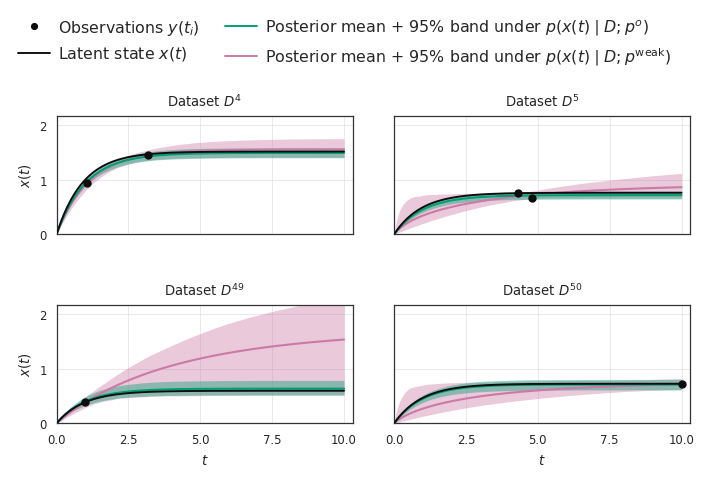}
\caption{LTI datasets: noisy measurements $y(t_i)$ (black circles), true state $x(t)$ (black line), and posterior mean trajectory with 95\% credible band under the true prior $p^o$, see Eq.~\eqref{eq:lti_generative} (green), and under the weak prior uniform in the range, see Eq.~\eqref{eq:lti_weak_interval} (pink).
}
\label{fig:lti_oracle_weak}
\end{figure}

The meta-dataset contains $\grs=50$ realizations. The first $48$ contain $\steps=2$ samples randomly spaced in the interval $[0, 10]$, while the last two (49\textsuperscript{th} and 50\textsuperscript{th}) feature a single measurement ($T=1$) at  instants $t_1^{49}=1$ and $t_1^{50}=10$, respectively. 
Fig.~\ref{fig:lti_oracle_weak} illustrates four representative datasets. For each of them, the measured points $y(t_i)$, the true state $x(t)$, and the posterior state mean---derived under knowledge of $p^o$---with 95\% credible bands are shown. 

We remark that, even for this simple example, the posterior lacks a closed-form analytical solution, and is here approximated with MCMC sampling. In this case, however, the MCMC implementation is simplified by the availability of the analytical solution of the ODE,
allowing the likelihood to be evaluated without numerical integration.
A comprehensive discussion of approximate probabilistic inference is deferred to Section~\ref{sec:implementation}.

It appears that, given knowledge of the generative prior $p^o$, the predictive uncertainty over the state remains remarkably narrow across all datasets. At the same time, if the prior $p^o$ was not known, some of the datasets in the collection would be very weakly informative. For instance, in $D^{5}$, the two data points are close to each other and both at steady-state, preventing practical identification of the dynamical parameters $\tau$ and $d$. Moreover, $D^{49}$ and $D^{50}$, which contain a single data point each, are structurally non-identifiable.

Now assume that the model structure~\eqref{eq:lti_structure} and constants $x_0=0, \sigma_\eta=0.05$ are known, while the generative distribution $p^o$ in ~\eqref{eq:lti_generative} is not. The only (correct, yet vague) background knowledge is that:
\begin{align}
\label{eq:lti_weak_interval}
\tau \in [0.1,\; 6.0], \qquad d \in [-6,\; 6]
\end{align}
i.e., that the system is stable, the time constant $\tau$ is not extremely small, and the absolute value of the steady-state $d$ is not extremely large. Based on this background, the Bayesian system identification practitioner may postulate a weak prior $p^{\mathrm{weak}}$,
uniform in the range given in Eq.~\eqref{eq:lti_weak_interval}\footnote{Alternative prior hypotheses compatible with~\eqref{eq:lti_weak_interval} are possible; the uniform over the interval is chosen for simplicity.}.
As shown in Fig.~\ref{fig:lti_oracle_weak}, under this weak prior the predictive uncertainty over the state trajectory becomes very large for $D^{5}$, $D^{49}$, and $D^{50}$, reflecting the difficulty of learning the dynamics without population-level information.

%% file: Content/21_MetaLearning.tex
\section{Probabilistic Meta-Learning of Dynamical Systems}
\label{sec:prob_meta_learning}
The shared structure assumption illustrated in Section~\ref{sec:shared_structure} is the key motivation for a meta-learning approach. Rather than instantiating independent learning problems for each dataset, where individual information may be insufficient, we instead devise a \emph{global} learning approach that explicitly takes advantage of this similarity across datasets. 
In principle, a probabilistic meta-learning approach aims to jointly estimate:
\begin{itemize}
    \item A shared prior distribution $\hat p(\theta)$ capturing the cross-dataset structure,
    \item The individual posterior distributions $p(\theta^{\gr} \mid D^{\gr};\; \hat p)$ for each dataset, constructed using $\hat p(\theta)$ as prior.
\end{itemize}
The interplay between prior learning and posterior inference enables the meta-dataset as a whole to inform parameter estimates for any single dataset.

\subsection{Hierarchical Bayesian Modeling}

The approach followed in this paper formalizes the prior learning objective by adopting an HBM framework. Fundamentally, HBM postulates that the meta-dataset is generated by the following hierarchical probabilistic process:
\begin{subequations}
\label{eq:HBM_framework}
\begin{itemize}
    \item Shared \emph{hyperparameters} $\hp$ are drawn from a \emph{hyperprior} distribution:
    \begin{equation}
        \label{eq:hyperprior}
        \hp \sim p(\hp).
    \end{equation}
    \item For each dataset $\gr \in \{1, \dots, \grs\}$, the system-specific parameters $\theta^{\gr}$ are drawn from a shared prior conditioned on those hyperparameters:
    \begin{equation}
        \label{eq:prior}
        \theta^{\gr} \sim p(\theta^{\gr} \mid \hp),
    \end{equation}
    with $p(\theta^{\gr} \mid \hp)$ being identical for all datasets.
    \item For each measurement instant $\step \in \{1, \dots, T_{\gr}\}$ within dataset $\gr$, the observation $y_{\step}^{\gr}$ is drawn from a measurement probability distribution conditioned on the specific system parameters and the sampling time $t_{\step}^{\gr}$:
    \begin{equation}
        \label{eq:lik}
        y_{\step}^{\gr} \sim p(y_{\step}^{\gr} \mid t_{\step}^{\gr}, \theta^{\gr}).
    \end{equation}
\end{itemize}
\end{subequations}

For systems of the form~\eqref{eq:system_equation}, the measurement
conditional density in~\eqref{eq:lik} is induced by the ODE solution and the observation model. In the case of known initial state and additive observation noise, it is given by 
\begin{subequations}
\begin{align}
    p(y_\step^\gr \mid t_\step^\gr, \theta^\gr)
    &=
    p_\eta(y_\step^\gr - \hat y_\step^\gr; \theta_\eta^\gr), \\
    \hat y_\step^\gr
    &=
    g(\hat x_\step^\gr, \theta_g^\gr), \\
    \hat x_\step^\gr
    &=
    x_0
    +
    \int_{0}^{t_\step^\gr}
    f(x^\gr(s),s,\theta_f^\gr)\,\mathrm{d}s .
    \label{eq:ode_solution}
\end{align}
\end{subequations}

The hierarchical formulation captures the shared structure across the meta-dataset. By modeling the parameters $\theta^{\gr}$ as originating from this shared prior $p(\theta^{\gr} \mid \hp)$, the datasets $D^\gr$ are structurally linked. Importantly, a vague hyperprior might be chosen in order to accommodate a wide range of plausible posterior scenarios.

The probabilistic model~\eqref{eq:HBM_framework} is then \emph{conditioned} on the observed meta-dataset to obtain the posterior distribution of hyperparameters and parameters of interest:
\begin{equation}
\label{eq:cond}
p(\hp,  \theta^{1:\grs} \mid \D) = 
\frac{p(\hp, \theta^{1:\grs}, \D)}{p(\D)},
\end{equation}
which describes the posterior belief of $\hp$ and $\theta^{1:\grs}$.

In general,~\eqref{eq:cond} does not have a closed-form expression, as its denominator $p(\D)$ is intractable. However, the \emph{joint} probability density function $p(\hp, \theta^{1:\grs}, \D)$ of hyperparameters, parameters and observations, i.e., 
\begin{equation}
\label{eq:joint}
p(\hp, \theta^{1:\grs}, \D) = p(\hp)\prod_{\gr=1}^{\grs} \bigg(p (\theta^{\gr} \mid \hp) \prod_{\step=1}^{\steps^\gr} p(y_\step^\gr \mid t_\step^\gr, \theta^\gr) \bigg),    
\end{equation}
for \emph{fixed} $\D$,
is proportional to the density of interest~\eqref{eq:cond}.  Therefore, an MCMC algorithm can be applied to obtain samples
from~\eqref{eq:cond}, using \eqref{eq:joint} as \emph{unnormalized} target distribution.

We denote the sequence of $M$ posterior samples (or trace) for both the shared hyperparameters and the system-specific parameters obtained with MCMC by:
\begin{equation}
\mathcal{T} = \Big\{\hp^{m}, \theta^{1;m}, \dots, \theta^{\grs;m} \Big\}_{m=1}^M \sim p(\hp, \theta^{1:\grs} \mid \D),
\end{equation}
where $\theta^{\gr;m}$ denotes the $m$-th posterior sample of the parameters associated with dataset $\gr$.
From the perspective of applied Bayesian analysis, the probabilistic model is considered~\emph{solved} once the trace $\mathcal{T}$ is computed, because these empirical samples characterize the otherwise intractable posterior distribution~\eqref{eq:cond}. Other predictive tasks, such as computing the posterior of a state trajectory $x^\gr(t)$, can be accomplished by forward-simulating the ODE for each parameter sample $\big\{\theta^{\gr;m}\big\}_{m=1}^M$ in the trace.

\begin{remark}[Non-additive Measurement Noise]
The additive noise assumption in Eq.~\eqref{eq:noise_on_state_equation} is introduced for notation simplicity, but it is not required by the Bayesian formulation. More generally, the observation model can be specified by an arbitrary conditional density
$y_i \mid t_i,\theta \sim p_Y\!(y_i \mid \hat y_i,\theta_\eta)$. 
For instance, for $y_\step \in \R^{n_y}_{+}$ a multiplicative log-scale noise model can be written
as $\log y_\step = \log \hat y_\step + \eta_\step$ with 
$\eta_\step \sim p_\eta(\cdot;\theta_\eta)$.
\end{remark}

\subsection*{Example: HBM for the LTI Meta-Dataset}
To apply the HBM framework to the LTI example, we first translate the deterministic belief~\eqref{eq:lti_weak_interval} on the plausible parameter ranges into a hierarchical probabilistic prior. We choose:
\begin{subequations}
\label{eq:lti_hyperprior}
\begin{alignat}{2}
    \tau^0 &\sim \N(3.05, 0.98^2),
    \qquad & \sigma_\tau &\sim |\N(0,2^2)| \\
    d^0 &\sim \N(0, 2^2),
    \qquad & \sigma_d &\sim |\N(0,2^2)|\\
    \tau^\gr &\sim \N(\tau^0, \sigma_\tau^2),
    \qquad & \gr&=1,\dots,\grs\\
    d^\gr &\sim \N(d^0, \sigma_d^2),
    \qquad &\gr&=1,\dots,\grs,
\end{alignat}
\end{subequations}
where $|\N(0,\sigma^2)|$ denotes a half-normal distribution with scale $\sigma^2$.
In~\eqref{eq:lti_hyperprior}, the coefficients induce prior distributions for $\tau^\gr$ and $d^\gr$ with a wide support that fully covers the ranges in~\eqref{eq:lti_weak_interval}.

The hierarchical prior~\eqref{eq:lti_hyperprior} is complemented with the Gaussian likelihood:
\begin{equation}
\label{eq:lti_lik}
y_\step^\gr \sim \N\big(d^\gr(1 - e^{-t/\tau^\gr}\big), 0.05^2),
\end{equation}
where $\gr=1,\dots,\grs$ and $\step=1,\dots,T^\gr$,
exploiting the closed-form solution of the first-order linear ODE with $x_0=0$ and the known form of the noise distribution.

\begin{figure}[htbp]
\centering
\includegraphics[width=\columnwidth]{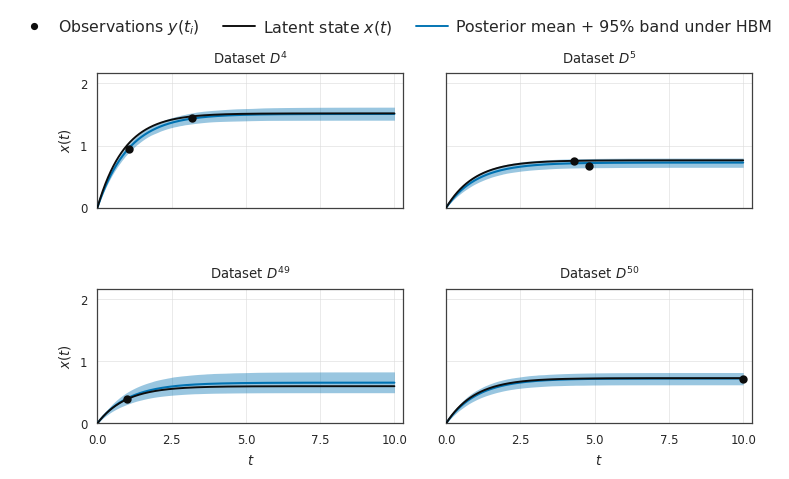}
\caption{LTI datasets: data points $y(t_i)$ (black circles), true state $x(t)$ (solid black line), posterior mean trajectory with 95\% credible band under the hierarchical model (blue).}
\label{fig:lti_hier}
\end{figure}

Overall, \eqref{eq:lti_hyperprior}-\eqref{eq:lti_lik} defines an HBM whose parameters can be estimated (i.e., conditioned on $\D$) with MCMC. 
We implemented the model in PyMC~\citep{abril-pla_pymc_2023} using NUTS and thus obtained a trace containing $M=10\,000$ {posterior} samples from:
\begin{equation}
    p(\tau^0, 
    \sigma_\tau, d^0, \sigma_d, \tau^{1:\grs}, d^{1:\grs} \mid \D).
\end{equation}
Fig.~\ref{fig:lti_hier} shows the posterior mean trajectories and 95\% credible bands of the hidden state $x(t)$ for datasets $D^4, D^5, D^{49}, D^{50}$.

We observe that the prediction quality is comparable to that obtained under knowledge of the true generative prior~\eqref{eq:lti_generative}, visualized in Fig.~\ref{fig:lti_oracle_weak}.

\subsection{Meta-Learned Prior for New Tasks}
To gain insights into the shared parameter structure learned by the HBM framework, it is instructive to visualize the distribution: 
\begin{equation}
\hat p(\theta^{\grs+1}) = p(\theta^{\grs+1} \mid \D),
\end{equation}
namely the posterior belief for the parameters of a~\emph{new}, unobserved dataset $D^{\grs + 1}$, assumed to share the same generation mechanism as $D^{1:\grs}$. We can identify this posterior predictive distribution as the \emph{learned prior}, since it can be used as an informative prior when performing inference on a future dataset. Formally, we can obtain $\hat p$ by marginalizing the \emph{conditional joint} 
$p(\hp, \theta^{\grs + 1} \mid \D)$:
\begin{equation}
\begin{aligned}
\hat p(\theta^{\grs+1})
&= p(\theta^{\grs + 1} \mid \D)
= \int p(\hp, \theta^{\grs+1} \mid \D)\; d\hp\\
&= \int p(\theta^{\grs+1} \mid \hp) p(\hp \mid \D)\; d\hp.
\end{aligned}
\end{equation}

In practice, given the sample-based approximation of $p(\hp \mid \D)$ obtained by solving the HBM with MCMC, samples of $\hat p(\theta^{\grs+1})$ can be obtained by sampling, for each $\hp^{m}$ in the trace, a corresponding ${\theta^{\grs+1; j}}$ from $p(\theta^{\grs+1} \mid \hp^{m})$. 

\begin{figure}[htbp]
    \centering
    \includegraphics[width=\columnwidth]{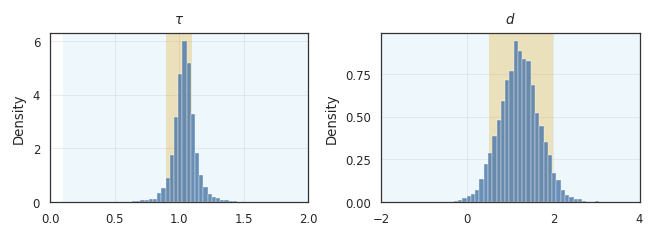}
    \caption{Histogram of $10\,000$ samples of the learned prior for $\tau$ (left) and $d$ (right). The uniform distribution corresponding to the ``true'' prior~\eqref{eq:lti_generative} is visualized as a superposed yellow area. The deterministic bounds~\eqref{eq:lti_weak_interval} corresponding to the prior knowledge are visualized as superposed blue shaded areas and extend beyond the horizontal limit of the figure. 
    }
    \label{fig:toy_lti_learned_posterior}
\end{figure}

Fig.~\ref{fig:toy_lti_learned_posterior} visualizes the learned prior $\hat p$ for the previous LTI example.
In particular, the learned prior for $\tau$ is strongly concentrated around the unit value, aligning with the underlying true generative distribution in~\eqref{eq:lti_generative}.
Given this fundamental additional information captured by the HBM through conditioning on $\D$, it becomes evident why joint estimation of both $\tau$ and $d$ is possible even in information-weak datasets like $D^{4}, D^{49}$, and $D^{50}$.

%% file: Content/22_Implementation.tex
\section{Implementation}\label{sec:implementation}

We recall that the practical implementation of MCMC requires repeated evaluations of the unnormalized target distribution, which in our case is the joint density~\eqref{eq:joint}.
Assuming the hyperprior, the dataset-specific prior densities, the measurement density,
and the functions \(f\) and \(g\) are inexpensive to evaluate, the primary technical difficulty in evaluating the joint distribution is the repeated solution of the ODE in~\eqref{eq:ode_solution}. In particular, when the analytical solution is unavailable, a numerical ODE integration scheme must be applied.
From a software implementation perspective, obtaining the conditional distribution~\eqref{eq:cond} for general systems of the form~\eqref{eq:system_equation} requires the coupling of a probabilistic inference engine implementing MCMC with a numerical ODE solver.

A key practical consideration in this coupling is whether the chosen ODE solver provides derivatives of the simulated state variables with respect to its parameters.
A differentiable solver can provide not only the simulated state $\hat x_\step^\gr$, but also the gradients of $\hat x_\step^\gr$ with respect to $\theta^\gr$, for example via a discrete or continuous adjoint method, as implemented in JAX-based solvers such as Diffrax~\citep{bradbury_jax_2018, kidger_on_2021}.

If such derivatives are available, the likelihood is differentiable with respect to $\theta^\gr$ through the full map
\begin{equation}
\theta^\gr \mapsto \hat x_\step^\gr \mapsto \hat y_\step^\gr \mapsto p(y_\step^\gr \mid t_\step^\gr,\theta^\gr),
\end{equation}
defined by the ODE solution in~\eqref{eq:ode_solution}.
Access to these gradients allows for the use of gradient-based MCMC methods such as Hamiltonian Monte Carlo or NUTS, which are attractive in high-dimensional landscapes because they can explore such posteriors efficiently. This is particularly relevant in the hierarchical setting, where the dimension of the posterior distribution grows with the number of datasets through the dataset-specific parameters $\theta^{1:\grs}$, in addition to the shared hyperparameters $\hp$. 

If the numerical solver is instead treated as a black-box routine, the joint density~\eqref{eq:joint} can still be evaluated, but gradients with respect to $\theta^\gr$ and $\hp$ are not available. 
This is the case, for instance, when standard solver implementations, such as SciPy's \texttt{solve\_ivp} are used.
Inference must then rely on gradient-free samplers, such as Metropolis-Hastings or slice sampling. While these methods may be adequate for low-dimensional or relatively simple models, they tend to become inefficient or numerically unstable for nonlinear ODE systems with strong posterior correlations across datasets.

In this work, we adopt a JAX-based implementation, using NumPyro~\citep{bingham_pyro_2019, phan_composable_2019} for probabilistic inference together with Diffrax for ODE solving. We use a 5th-order Runge-Kutta solver to numerically solve the ODEs, together with a PID step-size controller.
In our setup, the numerical integration step is part of the JAX computational graph, so that derivatives of the simulated states with respect to the sampled parameters are available to the gradient-based NUTS sampler.
The code is available online\footnote{\url{https://gitlab-core.supsi.ch/dti-idsia/hierarchical_bayesian_odes}}.
The experiments reported in the following section were run on a server equipped with two AMD EPYC 7742 64-Core processors (128 physical cores, 256 logical threads) and 629 GiB of RAM.

%% file: Content/30_Experiment.tex
\section{Numerical Experiments}\label{sec:experiment}

As a case study, we consider the Lotka–Volterra predator–prey model, a classical description of oscillatory population cycles in ecology.
We use parameter values representative of the classical Canadian lynx and snowshoe hare setting~\citep{stenseth_population_1997,carpenter_predator-prey_2018,brunkhorst_ode_2023} and simulate multiple datasets that we interpret as scenarios arising from different, but related ecological systems.
For example, these could correspond to distinct regions with varying environmental conditions.

\subsection{System Description}
The Lotka-Volterra model describes the interaction between predator and prey populations as:
\begin{equation}\label{eq:lotka_volterra}
    \dt x  = \dt
    \begin{bmatrix}
        x_1\\ 
        x_2
    \end{bmatrix}
    =
    \begin{bmatrix}
        \alpha x_1 - \beta x_1 x_2 \\
        \delta x_1 x_2 - \gamma x_2
    \end{bmatrix}
    = f\big(x,t;\,\theta_f\big),
\end{equation}
where the states $x_1$, $x_2$ represent the prey and predator populations in thousands of individuals, respectively, and $\theta_f = [\alpha\; \beta\; \gamma\; \delta]^\top$.

At a finite set of irregularly-spaced sampling instants, noisy measurements of both states are collected, corrupted by a mean-preserving multiplicative log-normal noise:
\begin{equation}
\label{eq:lv_observation_model}
    \begin{bmatrix}
        y_1(t_\step)\\ 
        y_2(t_\step)
    \end{bmatrix}
    =
    \begin{bmatrix}
        x_1(t_\step)
        \exp\!\bigg(
            \sigma_{\eta,1}\epsilon_{1,\step}
            - \frac{1}{2}\sigma_{\eta,1}^2
        \bigg) \\[3mm]
        x_2(t_\step)
        \exp\!\bigg(
            \sigma_{\eta,2}\epsilon_{2,\step}
            - \frac{1}{2}\sigma_{\eta,2}^2
        \bigg)
    \end{bmatrix},
\end{equation}
with $\epsilon_{1,\step},\epsilon_{2,\step}\sim\mathcal{N}(0,1)$ and $\sigma_{\eta,1}, \sigma_{\eta,2}>0$ unknown noise parameters.

\begin{remark}[Identification challenges]
The system possesses two steady states: the trivial equilibrium
$(x_1^\ast,x_2^\ast)=(0,0)$ and a coexistence equilibrium
$
    (x_1^\ast,x_2^\ast)
    =
    \left(
    \frac{\gamma}{\delta},
    \frac{\alpha}{\beta}
    \right),
$
around which system trajectories exhibit closed oscillations.
The coexistence equilibrium depends on the ratios $\frac{\gamma}{\delta}$ and $\frac{\alpha}{\beta}$, rather than on the individual parameters, so different parameter combinations that preserve these ratios generate similar trajectories, highlighting a source of weak practical identifiability. 
Different parameter vectors that preserve these ratios can generate rather similar trajectories, so precise estimation requires sufficiently informative data~\citep{raue_structural_2009}.
In a setting of noisy, irregularly sampled, and sparse data, standard single-dataset estimation becomes sensitive to prior assumptions~\citep{raue_joining_2013}.
\end{remark}

\subsection{Meta-Dataset Generation}
We generate $\grs=20$ Lotka-Volterra systems and corresponding datasets
\begin{equation}\label{eq:meta_dataset_lv}
    \mathcal{D} = \{D^{1}, D^{2}, \ldots, D^{20}\}.
\end{equation}

In all datasets, we fix $x_0=[35\;4]^\top$, corresponding to 35 thousand prey (hares) and 4 thousand predators (lynx).
The ODE coefficients $\theta_f = [\alpha\; \beta\; \gamma\; \delta]^\top$, and the noise coefficients $\theta_\eta = [\sigma_{\eta,1}\; \sigma_{\eta,2}]^\top$ vary across system realizations and form the parameter vector $\theta = [\theta_f^\top\; \theta_\eta^\top]
\in \mathbb{R}^{n_\theta}$ with $n_\theta=6$.
Each parameter in each dataset is generated from a scaled Beta family:
\begin{equation}
\label{eq:lv_generative}
\theta_k^\gr
\sim
\lambda_{\min,k}
+
(\lambda_{\max,k}-\lambda_{\min,k}) \mathrm{Beta}(\kappa_k,\nu_k), 
\end{equation}
for $k=1,\dots,n_\theta$ and $\gr=1,\dots,\grs$. 
The hyperparameters of the scaled Beta distributions are reported in Table~\ref{tab:lv_generative_params}.

\begin{table}[htbp]
\centering
\caption{Parameter distributional values used in the dataset-generating process.}
\label{tab:lv_generative_params}
\begin{tabular}{c c c c c}
\toprule
Parameters & $\kappa$ & $\nu$ & $\lambda_{\min}$ & $\lambda_{\max}$ \\
\midrule
$\alpha$          & 2.5 & 2.5 & 0.420 & 0.600 \\
$\beta$           & 2.0 & 5.5 & 0.013 & 0.027 \\
$\gamma$          & 5.5 & 2.0 & 0.780 & 1.020 \\
$\delta$          & 4.0 & 2.5 & 0.018 & 0.032 \\
$\sigma_{\eta,1}$ & 2.3 & 2.3 & 0.200 & 0.420 \\
$\sigma_{\eta,2}$ & 2.3 & 2.3 & 0.220 & 0.460 \\
\bottomrule
\end{tabular}
\end{table}

For each dataset $D^j$, we collect $T=5$ noisy state samples irregularly spaced over the interval $[0,t_\mathrm{end}]$, where $t_\mathrm{end}$ is 27 years, corresponding to around 3 population cycles.
The sampling instants are randomized across datasets, increasing the diversity in the meta dataset.

\subsection{Probabilistic Meta-Learning with HBM}
We assume access to the meta-dataset $\D$ in~\eqref{eq:meta_dataset_lv}, knowledge of the functional form of the data-generating mechanism~\eqref{eq:lotka_volterra} and of the noise structure~\eqref{eq:lv_observation_model}. However, the values of the parameters $\theta$ for each dataset are unknown, as are their probabilistic structure~\eqref{eq:lv_generative} and the hyperparameters in Table~\ref{tab:lv_generative_params}.


Instead, we assume some weak background knowledge on \emph{plausible} parameter ranges, informed by~\citet{carpenter_predator-prey_2018}.
Upper and lower range constituting weak prior knowledge on $\theta_f$ are given in Table~\ref{tab:lv_weak_interval}.
\begin{table}
\centering
\caption{Weak prior knowledge on the system parameters $\theta_f$.}
\label{tab:lv_weak_interval}
\begin{tabular}{c r r}
\toprule
Parameters & Lower bound & Upper bound \\
\midrule
$\alpha$ & 0.20 & 0.85 \\
$\beta$ & 0.006 & 0.035 \\
$\gamma$ & 0.45 & 1.30 \\
$\delta$ & 0.012 & 0.038 \\
\bottomrule
\end{tabular}
\end{table}
Note that these intervals are wider than the ranges used to generate the datasets.
For the HBM parameter estimation, this weak information is encoded through a hierarchical prior on the dataset-specific parameters.
For each dataset $j$ and the ODE parameters $\theta_f^j$, we introduce a non-centered Gaussian hierarchy on an unconstrained variable,
\[
z_\theta^j=\mu_\theta + \tau_\theta\epsilon_\theta^j, \qquad
\epsilon_\theta^j\sim\mathcal N(0,1),
\]
where $\mu_\theta$ represents the population location and $\tau_\theta$ the between-dataset variability of parameter $\theta_f^j$ on the unconstrained scale.
We then map $z_\theta^j$ to the admissible interval using a logistic transform combined with a min-max scaling.
This construction keeps all dataset-specific parameters within their weakly plausible ranges while allowing their population location and variability to be learned from the meta-dataset.
The population-level hyperparameters $\mu_\theta$ and $\tau_\theta$ are assigned weakly informative priors,
\[
\mu_\theta \sim \mathcal{N}(0,1),\qquad
\tau_\theta \sim \lvert\mathcal{N}(0,(2/3)^2)\rvert,
\]
for all four parameters.

For the weak prior knowledge on $\theta_\eta$, lognormal priors centered at the nominal values $\bar\sigma_{\eta,1}=0.30$ and $\bar\sigma_{\eta,2}=0.34$, are assigned, with half-normal priors on the corresponding between-dataset log-scale variation. 
The HBM was sampled with four NUTS chains, using 2000 warm-up iterations and 2500 posterior draws per chain, yielding a trace $\mathcal{T}$ of $M = 10\,000$ posterior samples.\footnote{Convergence diagnostics ($\hat{R}$ statistics, effective sample sizes, and trace plots) are provided in the code.} 

As a reference, we consider two models that estimate the parameters separately for each dataset $D^j$, without exploiting the shared structure of the meta-dataset.
As a Bayesian baseline, we consider an unpooled model that treats the datasets as independent. It uses the same likelihood, parameter transformations, and sampling setup as the HBM described above, but assigns independent bounded weak priors to the transformed ODE parameters and independent priors to the two dataset-specific observation-noise scales, without any shared population-level hierarchy.

As a non-probabilistic baseline, we compute nonlinear least-squares (NLS) estimates with box constraints over the weak ranges in Table~\ref{tab:lv_weak_interval}. We use a multi-start optimization strategy and residuals defined as differences between the logarithms of simulated and observed states, and return a single point estimate for the four ODE parameters $\theta_f$ in each dataset, without estimating observation-noise parameters $\theta_\eta$.

We compare the performance of the three estimation approaches with respect to both the accuracy of the trajectories and the parameter recovery.
Trajectory accuracy is assessed on a fine grid of 1000 points by comparing estimated state trajectories with the true ones.
For NLS, we obtain a point estimate of the state trajectory by simulating~\eqref{eq:lotka_volterra} with the obtained estimate of $\theta_f$.
For the Bayesian models, we use as point estimate the posterior mean, computed as the point-wise mean over trajectories obtained by simulating~\eqref{eq:lotka_volterra} with parameters 
from the MCMC posterior trace.
We report average root mean squared error (RMSE) values for the trajectories. 
Furthermore, for the Bayesian models we assess the latent-trajectory uncertainty using the latent-state continuous ranked probability score (CRPS)~\citep{gneiting_strictly_2007}, averaged over time on the fine grid.

For parameter estimation, we first compute the relative error between the true dataset-specific parameters and the corresponding point estimates. We summarize overall recovery accuracy as a normalized RMSE (nRMSE), computed for each dataset as the root mean squared value of these relative errors across parameters, and then averaged across datasets. For the Bayesian models, point estimates of the parameters are defined as the posterior mean of the parameters. To quantify parameter uncertainty, we also report the mean relative width of the corresponding 95\% highest density intervals (HDI).

\subsection{Results}

\begin{figure}[htbp]
\centering
\includegraphics[width=\columnwidth]{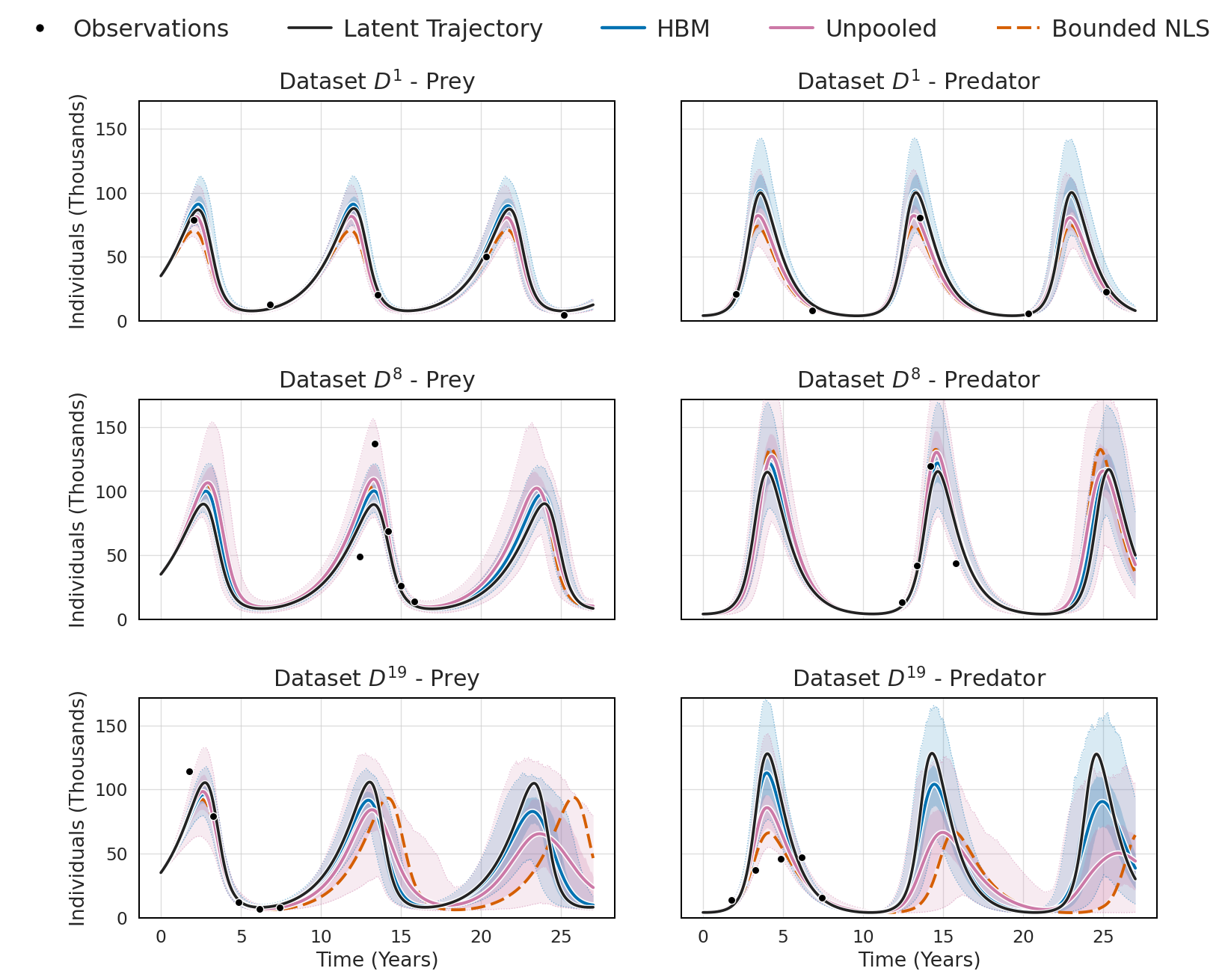}
\caption{Latent trajectory estimates for three representative Lotka--Volterra datasets. Black circles denote the noisy observations, solid black curves the true latent states for prey and predators, blue curves the HBM posterior mean trajectories, pink curves the unpooled Bayesian posterior mean trajectories, and orange dashed curves the bounded NLS fits. Shaded bands indicate 50\% and 95\% posterior credible intervals for the Bayesian models.}
\label{fig:lv_trajectory_comparison}
\end{figure}

The complete HBM run required 20 minutes on the computing setup described in Section~\ref{sec:implementation}.
In terms of mean trajectory RMSE, the HBM model achieves better performance (RMSE of 8.26, in the units of the ODE states), compared the unpooled Bayesian model (11.77) and the NLS (17.17).
This represents a substantial improvement in trajectory accuracy for the hierarchical approach, corresponding to a reduction of approximately 30\% over the unpooled Bayesian model and more than 50\% over NLS. 
Fig.~\ref{fig:lv_trajectory_comparison} visualizes the latent trajectory estimates for three representative datasets $D^1$, $D^8$, $D^{19}$.
From the figure, it is evident that the performance of the three methods depends on the informativeness of each dataset.
For $ D^1$, where the observations are well-positioned to constrain the dynamics, both the HBM and the unpooled model recover the true trajectory accurately, although the HBM already achieves a better fit for the predator state. For dataset $D^8$, the HBM starts to outperform both baseline models more clearly, and for $D^{19}$, where the sparse and irregularly sampled observations provide limited information, the performance gain is evident.
This improvement is also reflected in the distributional trajectory score. The HBM achieves a mean latent CRPS of 3.59, compared to 5.18 for the unpooled model.

For parameter recovery, the HBM is more accurate (parameter nRMSE of 0.07) compared to the unpooled model (0.12) and the NLS (0.20).
The mean and standard deviation of the relative error for each parameter are reported in Tab.~\ref{tb:LV_estimated_parameters}.
\begin{table}[htbp]
\centering
\caption{Mean relative bias and standard deviation (in brackets) of parameter estimates across the 20 datasets.}\label{tb:LV_estimated_parameters}
\small
\setlength{\tabcolsep}{4pt}
\begin{tabular}{lccc}
\toprule
 & HBM & Unpooled & NLS \\
\midrule
$\alpha$ & 0.02 (0.05) & 0.05 (0.07) & 0.02 (0.14) \\
$\beta$  & 0.02 (0.10) & 0.13 (0.18) & 0.11 (0.37) \\
$\gamma$ & -0.01 (0.04) & -0.03 (0.07) & -0.01 (0.13) \\
$\delta$ & -0.01 (0.08) & -0.05 (0.09) & -0.03 (0.18) \\
\bottomrule
\end{tabular}
\end{table}
Across all four parameters, the HBM exhibits consistently lower bias and notably lower variability across datasets.
The posterior uncertainty, quantified by the mean relative width of the 95\% HDI, also demonstrates the benefit of hierarchical pooling. The HBM achieves a mean relative HDI width of 0.35, compared to 0.58 for the unpooled model.
This indicates that the HBM produces credible intervals that are approximately 40\% narrower on average. Importantly, because this reduction in uncertainty is accompanied by lower trajectory RMSE, lower CRPS, and improved parameter recovery, the results indicate that the HBM is not simply overconfident but provides more accurate and well-calibrated estimates.

%% file: Content/40_Conclusion.tex
\section{Conclusions}\label{sec:conclusions}
We introduced a hierarchical Bayesian approach to probabilistic meta-learning for dynamical systems, enabling joint inference across related datasets.
By learning a shared prior distribution from the meta-dataset, the proposed approach mitigates practical non-identifiability arising in sparse and irregularly sampled settings, where single-dataset inference is unreliable.
The integration of differentiable ODE solvers with gradient-based MCMC allows efficient numerical inference, for which we provide an open-access implementation to facilitate adoption. 
Numerical results on a Lotka--Volterra benchmark demonstrate that the method consistently outperforms standard unpooled Bayesian and nonlinear least-squares approaches.
This highlights the potential for data-efficient system identification in applications with sparse and noisy data, such as those arising in the biomedical domain.
As a promising direction, the framework could be extended to hybrid gray-box models that combine dynamical systems with Bayesian neural network components for increased expressivity.